# An Artificial Life Simulation Library Based on Genetic Algorithm, 3-Character Genetic Code and Biological Hierarchy

**Maurice HT Ling**
*Department of Mathematics and Statistics*
*South Dakota State University, USA*
*Department of Zoology*
*The University of Melbourne, Australia*
mauriceling@acm.org

## Abstract

Genetic algorithm (GA) is inspired by biological evolution of genetic organisms by optimizing the genotypic combinations encoded within each individual with the help of evolutionary operators, suggesting that GA may be a suitable model for studying real-life evolutionary processes. This paper describes the design of a Python library for artificial life simulation, Digital Organism Simulation Environment (DOSE), based on GA and biological hierarchy starting from genetic sequence to population. A 3-character instruction set that does not take any operand is introduced as genetic code for digital organism. This mimics the 3-nucleotide codon structure in naturally occurring DNA. In addition, the context of a 3-dimensional world composing of ecological cells is introduced to simulate a physical ecosystem. Using DOSE, an experiment to examine the changes in genetic sequences with respect to mutation rates is presented.

## 1. Introduction

Life can be viewed as an optimization to the surrounding environment. The atomic unit for life is a cell. Central to a cell is the genetic code, which can be visualized as a complex set of instructions for the cell to interact with the environment. A large number of biological research into evolution had demonstrated that mutation or changes in the genetic code occurs as the organism in question adapts to a new environment (Cooper et al., 2008) or gains new properties (Goh et al., 2012; How et al., 2012). These may include resistance to drugs and medical treatments (Bibbal et al., 2009), or adaptation to different chemicals (Lee et al., 2012) and temperatures (Tosun and Gonul, 2005). These have inspired a class of heuristics known as genetic algorithms (GA).

GA is where the solution of a problem is encoded into a string (Engelbrecht, 2007), known as a chromosome. These are operated on, using biologically equivalent operations, such as mutations and translocations. A cell can have one or more chromosomes, forming a genome. The process of mating in sexual organisms is then an assortment and combination of the genetic information in the chromosome to form a new generation. This is known as the law of independent assortment or the inheritance law in classical Mendelian genetics. GA had been used successfully in a number of applications (see Shiekh et al. (2008) for a review), including simulation of evolutionary processes of biological genomes (Dalquen et al., 2011).

The biggest issue in studying evolution is time. The generation time for most multicellular organisms can range from days (such as insects) to years (such as humans). Even using fast-growing bacterium, such as *Escherichia coli*, only a small number of generations can be studied in a day. Recent publications suggest about 7 generations of *Escherichia coli* can be achieved in standard laboratory conditions (Goh et al., 2012; Lee et al., 2012). On the other hand, modern computers are able to simulate thousands of generations within hours. Hence, there is a significant time advantage in studying evolutionary processes *in silico*.

Christopher Langton (1986) had conceptualized that by casting chemical reactions, reactants, and products into computable operations, operands, and outputs respectively, it may be possible to simulate artificial life (organisms in the digital world, or digital organisms) as cellular automata "living" on artificial chemistries. Thus, the field of artificial life (ALife) is created and had been used in many different domains (Ward et al., 2011; Kim and Cho, 2006). In the field of biology, Bersini (2009) argued that artificial life and theoretical biology shared many common grounds and presented GA as an important model to bridge the two fields. GA was used to study microbial genetics and evolution (Harvey, 2011). This corroborates another study arguing that evolution by natural selection is the algorithm of biological evolution (Watson, 2012).

A number of ALife simulators had been developed over the years (Bornhofen and Lattard, 2006; Komosinski and Adamatzky, 2009). Common ALife simulators include Tierra (Ray, 1992), Echo (Holland, 1992), Polyworld (Yaeger, 1994), Framesticks (Komosinski, and Ulatowski, 1999), Avida (Ofria and Wilke, 2004), and EcoSim (Gras et al., 2009). Recently, a Python GA framework conforming to biological hierarchy starting from gene to chromosome to genome (as organism) to population had been developed (Lim et al., 2010), which may help interpreting GA results to biological context.

In this study, the GA framework (Lim et al., 2010) is expanded into a digital organism simulation environment (DOSE) by formalizing a 3-character genetic language to correspond the codon (which comprises of 3 nucleotides) in naturally occurring DNA and incorporating a 3-dimensional "world". The organization of this paper is as follows: Section 2 presents an architectural description of DOSE and its biological context. Section 3 discusses the artificial chemistry of DOSE. Section 4 presents two experiments using DOSE to illustrate its use. Section 5 concludes this paper by discussion on the strengths and limitations of DOSE and its future work.

## 2. DOSE Architecture

DOSE is based on the hierarchical structure of Lim et al. (2010) and designed as a library for importation into a simulation driver program. Data propagates from the world, at the highest level, to the chromosome in each organism at the lowest level where the genomic instructions are interpreted and the results propagates back to the world via organism and population layers (Figure 1). Another example of an ALife simulator that uses a hierarchical architecture is that of Curran and O'Riordan (2003).

DOSE comprises of 3 distinct components – a GA framework (Lim et al., 2010) providing for chromosome, genome, organism and population hierarchy, a 3-dimensional world in which the population exist within, and a set of 3-character known as Ragaraja instruction set which acts as genetic code and an interpreter to execute Ragaraja instruction. Hence, at the core of DOSE is executable DNA, which is a common design, used in a number of ALife simulators including Avida (Ofria and Wilke, 2004).

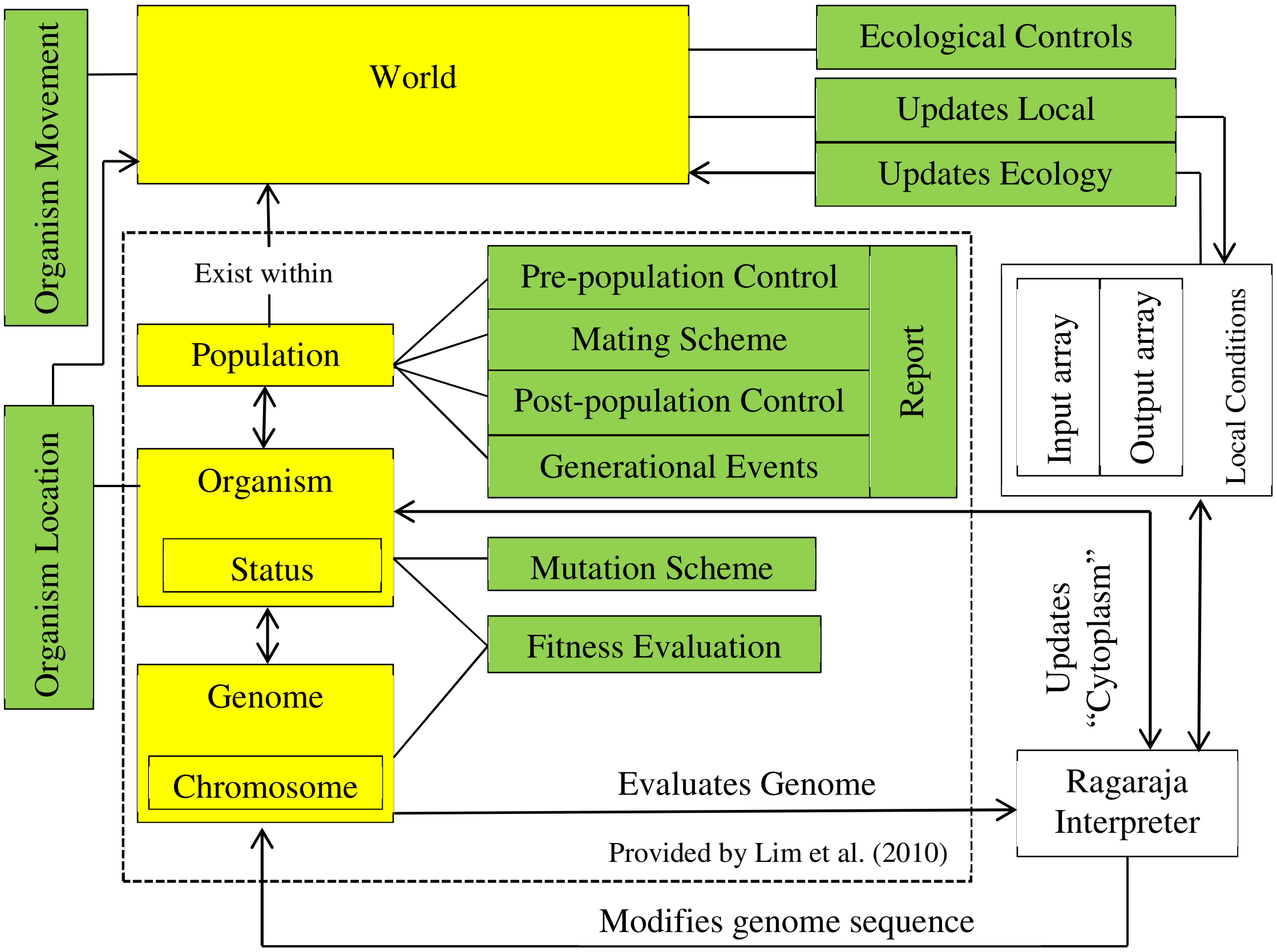

**Figure 1. Architecture of DOSE.** The yellow boxes represent the entities while the green boxes represent user-defined functions as point of control that regulates various aspects of the digital organism and the ecology in which the organisms are in. Table 1 provides a summary of the user-defined functions and examples of its use.

In the GA framework (Lim et al., 2010), the most important object is an *Organism*. The status of an organism is logged within its *status* dictionary and consists of one chromosome as genome. It is not essential for an organism to have a genome. In the case of a genome-less organism, the fitness function, provided by the user, will evaluate the fitness of the organism based on its status. In an organism with a genome, the user can provide a mutation scheme that will be activated once per generation by default. The concept of an organism is unicellular. A population can simulate a multicellular organism. At the population level, the user can provide a number of functions, such as pre-population control, to manage the population before mating; mating scheme, for mate choices and reproduction; post-population control, to manage the population after mating; and generational events, as a catch-all for any other events defined by the user. For example, one mutational event per generation default can be over-rode at this step by calling *Organism.mutation_scheme* function once or more times. At each generation, the user can define a report function to report the fitness and conditions of the population.

Naturally occurring DNA comprises of 4 nucleotides – adenine, thymine, guanine, and cytosine – commonly abbreviated as “A”, “T”, “G”, and “C” respectively. A protein chain is made up of 20 possible amino acids. Using only 4 nucleotides to code for 20 possible amino acids, the nucleotides are read in triplets (known as codon) which results in 64 ($4^3$) different codons. The resultant protein chain will fold into a 3-dimensional structure, known as a tertiary structure, and may bind with other protein chain to form a quaternary structure. These structures determine the activity of the proteins. The site of activity is known as a protein domain. Using these biological concepts as basis, Ragaraja[1] is an esoteric programming language comprising of a set of 3-character instructions to mimic a codon. As a derivative of BrainFuck, Ragaraja uses 3 numbers as an instruction instead of symbols of alphabets. This simplified the implementation of a mutation scheme. Conceptually, a Ragaraja-encoded genome may be visualized as having 10 nucleotides instead of 4. Thus, there are a total of 1000 possible instructions in Rajaraga but only 347 are defined and used in this current version, Version 1 (see Appendix A for description of each instruction). In addition, all instruction are atomic and do not take on any operand which mimics natural DNA. However, there is no equivalence of a 3-dimensional protein structure in Ragaraja. Despite so, each Ragaraja instruction is an operation and in biological terms, an operation can be deemed as a protein domain. Hence, it can be seen that protein domains are encoded directly in the genome of DOSE organism and there is a possible to define up to 1000 protein domains in the current 3 number encoding.

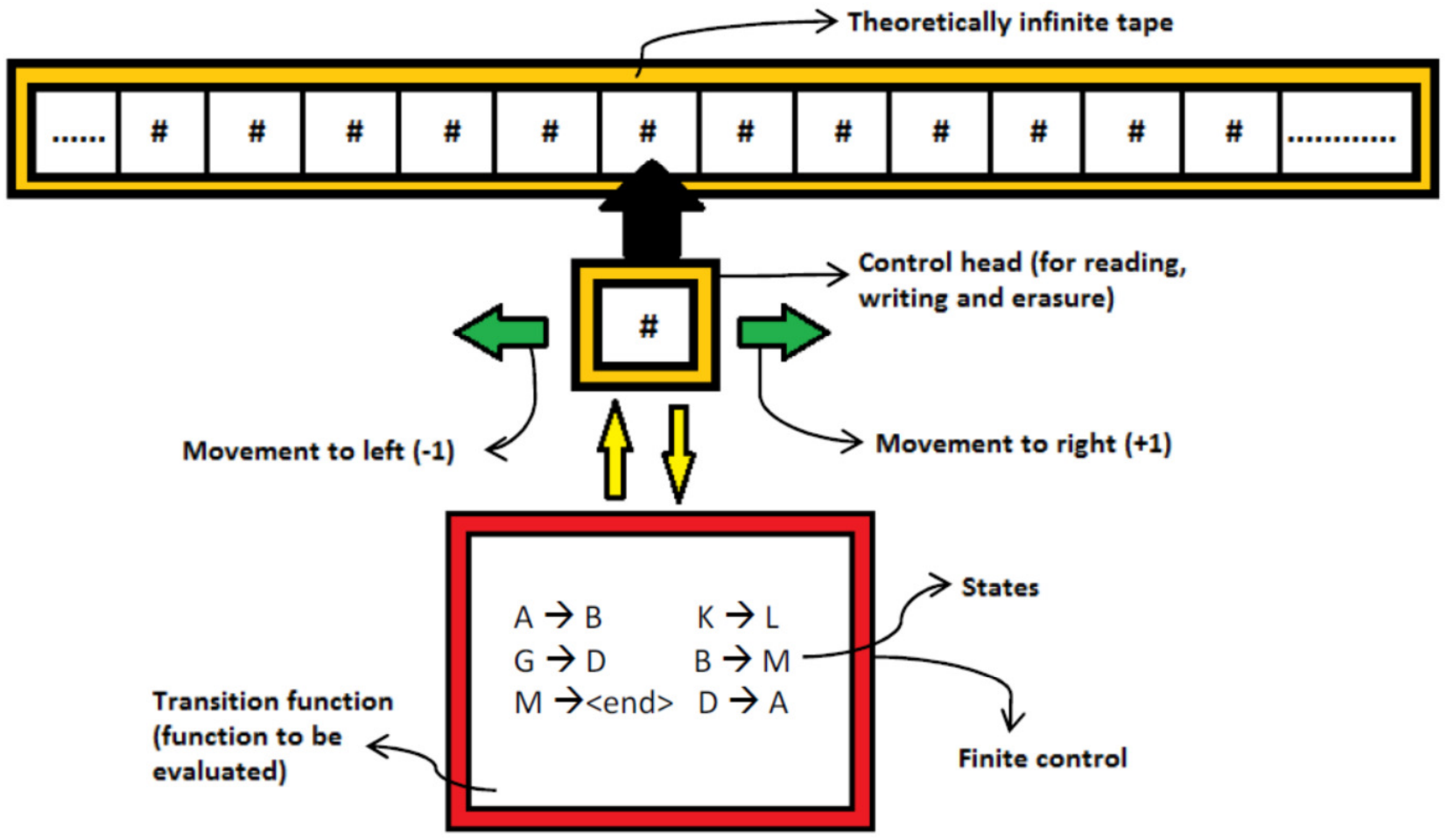

**Figure 2. Conceptual scheme of a Turing machine.** The transition function can be visualized as the function to be computed or a set of rules for execution. Adapted from (Pereira et al., 2010).

BrainFuck is a Turing complete language (Kohler et al., 2005; Ling, 2010a); thus, able to implement any computable functions. By extension, a Turing machine is a machine capable of performing any computable functions. It had been suggested that Turing complete language and a Turing machine are essential for ALife simulation (Langton, 1986). Rajaraga includes the 7 computing operations of BrainFuck in its definition. Therefore, Rajaraga is a Turing complete language by virtue of extension.

[1] http://esolangs.org/wiki/Ragaraja

One of the most enduring images of a Turing machine (Figure 2) is a virtually infinite tape with a set of rules to be executed based on the information on the location of the read/write head on the tape (Pereira et al., 2010). Ragaraja interpreter uses 4 tapes – a source/instruction tape to contain the operations to be executed, an event tape to represent the theoretically infinite tape where computations are carried out, an input tape to fed input into the system, and an output tape to collect any output from the execution. In addition, Ragaraja interpreter defined a set of 99 general-purpose registers.

Putting into biological context, the source/instruction tape will be equivalent to the genome. The event tape will be similar to the cytoplasm of a cell where all reactions take place. The set of 99 registers may be used to simulate the presence of compartments within the cell, such as organelles. However, these registers are not available for fitness evaluation. The input and output tapes may collectively represent the extracellular matrix of the cell. More specifically, the input and output tapes can represent the endocytotic (molecules taken into a cell) and secreted components respectively. In the context of DOSE, each genome is sent for execution and the results are evaluated by user-defined fitness function per generation or time lapse.

The concept of a world is defined as a 3-dimensional cellular automaton model. Therefore, each non-edged ecological cell has 26 neighbouring ecological cells – 6 full-face contact adjacent ecological cells (front, back, left, right, top, and bottom), 8 diagonal edge contact ecological cells (top-front, top-back, top-left, top-right, bottom-front, bottom-back, bottom-left, and bottom-right), and 12 diagonal non-edge contact ecological cells. The term "ecological cell" is used to describe each cell in this 3-dimensional cellular automation world to prevent confusion from a "biological cell". Each organism is mapped onto an ecological cell and is aware of its location in the ecosystem. Thus, each cell can have one or more organisms. As a result, a population of organisms can occupy one or more ecological cells. Although mating should only occur within an ecological cell, there is no restriction as cross-cell mating can be defined in the population mating (*Population.mating* function) scheme. Each ecological cell has a set of uniformed local conditions and a collection of ecological cells forms an ecosystem.

On the world-scale, the ecosystem is controlled or regulated by user-defined ecological controls. The entry point of these controls is *World.ecoregulate* function where the user can define one or more controlling functions, depending on the experiment. The local conditions of each ecological cell are determined by the ecosystem and the input/output of each organism within the cell. Local conditions of each ecological cell can affect the ecosystem at large. The role of *World.update_ecology* function, given by the user, includes normalizing the local ecological cell condition from the input/output of each organism within the cell and reflecting these changes to the ecosystem. As the ecosystem is not an entity but a collection of ecological cells, the *World.update_ecology* function acts to radiate local conditions to adjacent or nearby cells. For example, a user can use this function to define chemical events such as pheromone diffusion or heat radiation from a local ecological cell. Conversely, the ecosystem can affect the local condition and such effect is defined by *World.update_local* function, which is given by the user. A common scenario for this use is temperature gradients across the world. *World.update_local* function may also be used to trigger regional mutation events in the organisms.

| Functions | Usage | Examples of Use |
|---|---|---|
| O.fitness | Calculates the fitness of the organism and returns a fitness score. | Determining genetic fitness of an organism. |
| O.mutation_scheme | Trigger mutation events in each chromosome. Helper functions are *Chromosome.rmutate* (for random mutation throughout the genome) and *Chromosome.kmutate* (specific mutation within a segment of the genome). | Simulate mutation events. |
| P.prepopulation_control | Trigger population control events before mating event in each generation. | Simulates pre-puberty (childhood) death. |
| P.mating | Trigger mating events in each generation. Helper function is *crossover* function in Lim et al. (2010) | Simulates mate choices and progeny size. |
| P.postpopulation_control | Trigger population control events after mating event in each generation. | Simulates old-age death. |
| P.generation_events | Trigger other defined events in each generation. | Simulates catastrophe or epidemic that does not occur regularly.<br>Simulates unusual occurrences of multiple mutation events. |
| P.report | Report the status of each generation. | Produce output for analyses between generations. |
| W.organism_movement | Movement of organisms within the world. | Simulates foraging or nomadic behaviour. |
| W.organism_location | | Simulates long distance migration, such as air travel. |
| W.ecoregulate | Simulate events to the entire ecosystem. | Simulates temperature and resource gradients. |
| W.update_ecology | Process the input and output from the activities of the organisms in the current ecological cell into a local ecological cell condition, and update the ecosystem. | Simulates secretion of chemicals or use of resources (such as food) by organisms, and diffusion of secretions to the neighbouring ecological cells. |
| W.update_local | Update local ecological cell condition from the ecosystem. | Simulates movement or diffusion of resources from the ecosystem to local. |
| W.report | Report status of the world (ecosystem) | Produce output for analyses between generations. |

**Table 1. Summary of user-defined functions.** "O" in *O.fitness* function represents organism. "P" represents population. "W" represents world.

DOSE world support 2 additional functions, which can be used to simulate the movement of organisms within the world – *World.organism_movement* and *World.organism_location* functions. The *World.organism_movement* can be used to simulate migration based on immediate and neighbour ecological cell location. A possible use may be to simulate an organism in search of greener pastures or dealing with over-population or under-population.

The *World.organism_location* can be used for long distance migration. Table 1 provides a summary of the 13 user-defined functions and examples of its use.

However, this does not mean that it is essential to use the world as part of the simulation. In the simplest scenario, the world can be defined as just a single ecological cell and the entire population exist within a single ecological cell. In this case, only the *World.update_ecology* function needs to be defined to normalizing the local cell condition from the input/output of each organism.

In order to execute the simulation, the following default simulation driver is provide to execute the simulation, given one or more pre-allocated populations in the world:

```
Run World.ecoregulate function
For each ecological cell
     Run World.update_ecology function
     Run World.update_local function
For each organism
     Execute genome by Ragaraja interpreter using
          existing cytoplasm, local conditions as input
     Update cytoplasm (Organism.cytoplasm)
     Add input/output from organism temporary conditions
          of local cell
For each population
     Run Population.prepopulation_control function
     Run Population.mating function and
          add new organisms to cell
     For each organism, run Organism.mutation_scheme function
     Run Population.generation_events function
     Add 1 to generation count
     Run Population.report function
     Fossilize population if needed (save into file)
For each ecological cell
     Run World.organism_movement function
     Run World.organism_location function
     Run World.report function
Bury ecosystem if needed (save into file)
Repeat simulation until maximum_generation is reached
```

## 3. Artificial Chemistry

Biochemistry is the study of the molecular basis of life whereby each individual component of a biochemical reaction is inanimate but together, forms a set of self-sustaining and self-regulatory chemical system, which appears more than the sum of its inanimate parts. Philosophers thought that living organisms are endowed with a divine life-force to “organic-ize” its inorganic chemistry. This doctrine, which came to be known as *vitalism*, had been rejected by modern science when Friedrich Wohler synthesized urea from silver isocyanate and ammonium chloride in 1828; thus, producing an organic compound (urea) without a kidney even though the term "organic chemistry" remain in use today. On the basis that biochemistry is a set of inanimate chemical reactions, Christopher Langton (1986) defined artificial chemistry as a man-made system of interactions between artificial molecules, which can be states on a Turing machine and the rules for changing the states are equivalent to artificial chemical reactions. Formally, Dittrich et al. (2001) defined artificial chemistry as a triple of ($S$, $I$) where $S$ is the set of possible molecules, and $I$ is the set of interactions or reactions between molecules.

There are different levels of artificial chemistry in DOSE. At the organism level, the artificial chemistry in DOSE is of machine-tape interaction as defined by Ikegami and Hashimoto (1995). In this setup, the set of tapes (cytoplasm and genome) and arrays (input and output) form the molecules while the genome forms the set of interactions in the form of:

$$Cytoplasm + Genome + Input + Output \xrightarrow{interaction} Cytoplasm' + Genome' + Input' + Output'$$

where the interaction is defined by each Ragaraja instruction. All Ragaraja instructions do not take any operand but defines the changes to the event tape (cytoplasm), source tape (genome), input and output arrays. There are a number of mathematical operations within Ragaraja instruction set. For example, instruction '010' adds the 10 to the value of the current cell in the event tape (cytoplasm). At this level, the artificial chemistry can be simply defined by arithmetic chemistry where the molecules are natural numbers and the reaction is a mathematical operation,

$$Cytoplasm[i] + 10 \xrightarrow{010} Cytoplasm[i]$$

However, chemical reactions exist within a context. For example, biochemistry exists within a watery solution where certain ions such as proton ($H^+$) or hydroxyl ions ($OH^-$) are assumed to be of unlimited supply. To cater to this, Dittrich et al. (2001) defined a third component in artificial chemistry – the reactor algorithm or dynamics ($A$), which is the rule or a set of rules applying to the collection of molecules. This defines the availability and concentrations of each molecule. In the context of DOSE, only molecules of limited supply is modelled by the input and output arrays. Therefore, each ecological cell can represent the availability of various molecules of limited supply. The concentrations of these molecules can be simply modelled as the relative quantities of these molecules. In addition, the spatial topology, flow and diffusion of such molecules across adjacent ecological cells in the world can be defined by *World.update_ecology* and *World.update_local* functions.

## 4. Experiment

An experiment was carried out to test the operations of DOSE. The biological significance is to examine the changes in genetic sequences with respect to mutation rates. Four point mutation rates were used (2%, 3%, 4%, and 5%) across the entire genome using a random mutation operator. Two populations were simulated. Each population consists of 100 organisms and the cytoplasm size is set between 50 and 200. The ancestor organism has a genetic sequence of "000000000000000000000000000000000000000000000000000000 0000" (60 zeros). In each generation, a clean cytoplasm is used. Using the 10th generation as

a standard, the genetic sequence distances at intervals of 10 generations were calculated as Hamming distances (Ling, 2010b). Code for the entities used in this experiment is given in Appendix E.

The results show that the Hamming distances are not proportional to the mutation rates (Figure 3). However, the Hamming distances at the same mutation rate are highly correlated with each other (r > 0.999). This is interesting as only random point mutations were used in this experiment; hence, a correlation between mutation rates and sequence divergence should be expected. However, results show no difference in average sequence divergence between 2% and 3% mutation rate but a noted difference between 3% and 4% mutation rate. This phenomenon had also been observed in nature (Ellegren, 2007). A possible explanation for this may be repeated mutation on the same nucleotide.

A.

B.

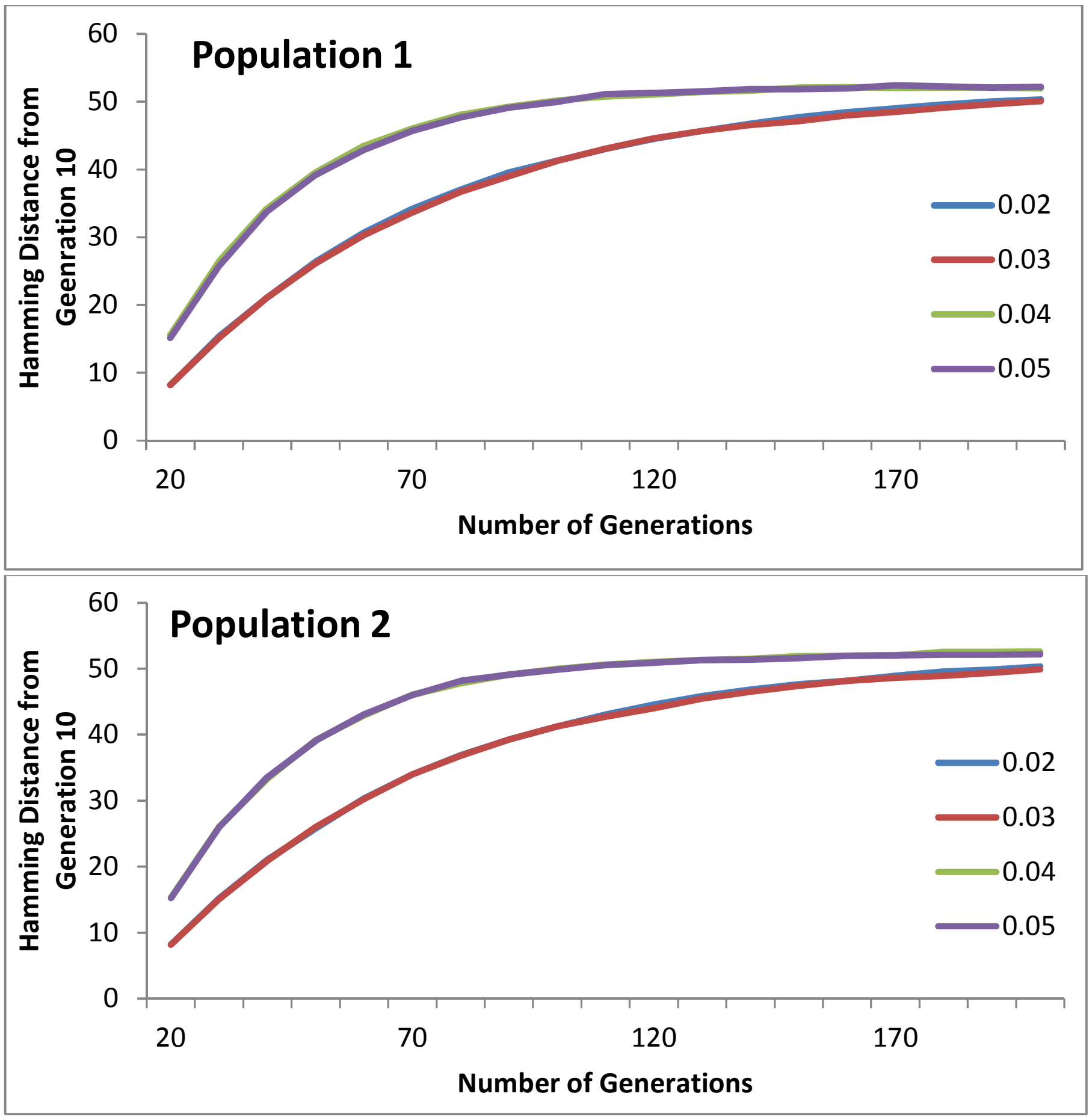

**Figure 3. Sequence distance across generations comparing with generation 10.** Panel A and B show data from Population 1 and Population 2 respectively. Four different mutation rates were used – 2%, 3%, 4%, and 5%. High correlations between the sequence distances of the same mutation rate in different population are observed (r > 0.999).

Using a subset of Ragaraja (NucleotideBF, see Appendix A) as the instruction set (Ragaraja version 0.1), cytoplasmic values were analysed. The results show that the average cytoplasmic value of the 100 organism approaches zero after about 80 generations and centred at zero for the rest of the simulation (Figure 4A). This is expected as there are only 2 increment and decrement operators in NucleotideBF. Hence, it can be expected that over the course of random mutations, the number of increments and decrements approaches the same number. However, there is no correlation between the average cytoplasmic values between the 2 populations (Figure 4C, $r^2$ = 0.003). This may suggest that the 2 populations are evolving independently from each other.

A. B. C.

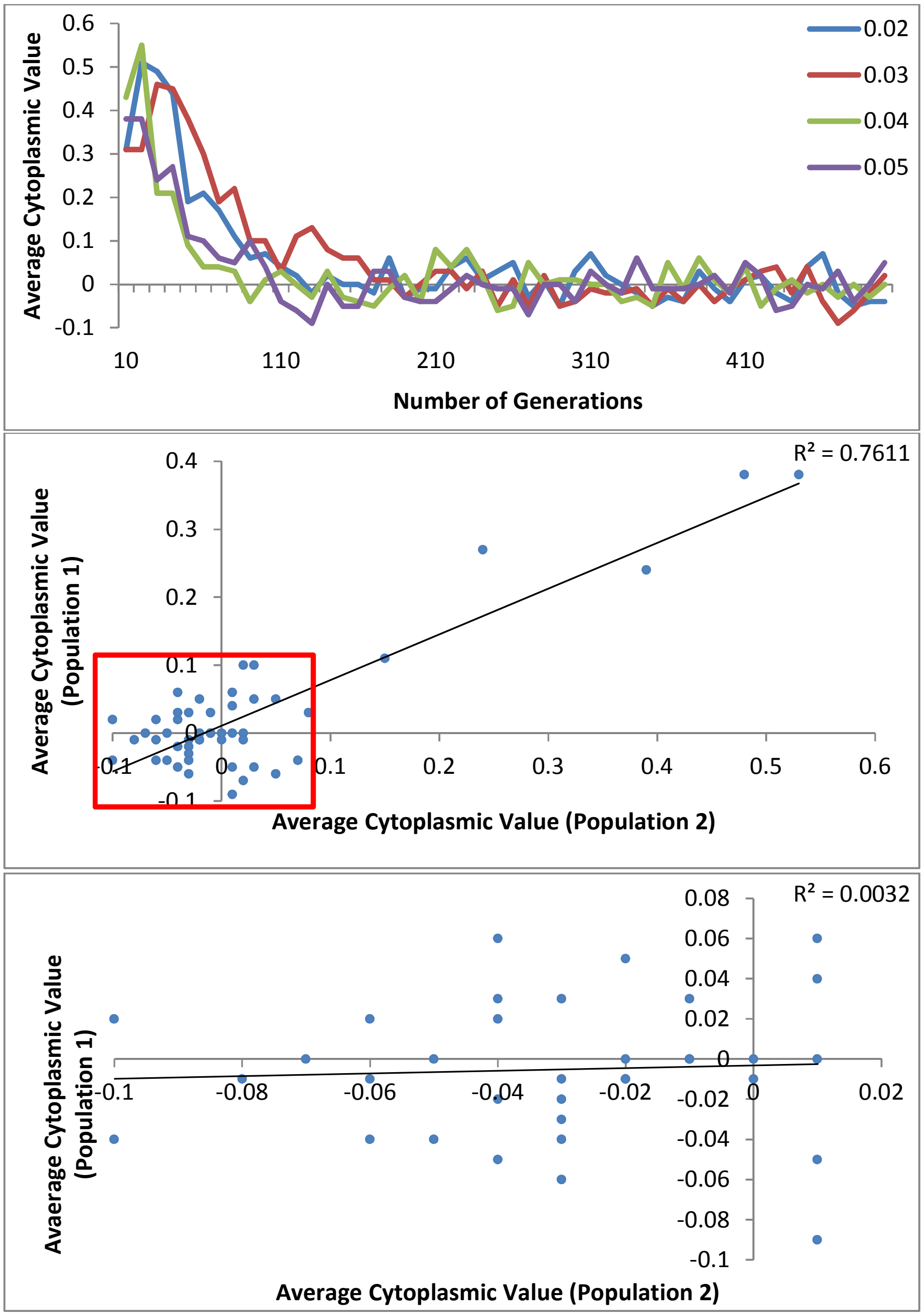

**Figure 4. Average cytoplasmic value across generations and the correlation of average cytoplasmic values between the 2 populations**. Panel A shows the average cytoplasmic value of Population 1 fluctuates around zero after generation 80 regardless of mutation rates. Panel B shows the correlation between average

cytoplasmic value of Population 1 and Population 2, paired by number of generations. Although a correlation coefficient is high ($r^2$ = 0.76), there is a clustering of points near to zero. As such, the values are not evenly distributed. Panel C expands on the near zero values of Panel B (in red box) and shows low correlation ($r^2$ = 0.003).

## 5. Discussion and Future Work

Resources and the generation time of the organisms of interest hinder the study of evolution on a biological platform. On the other hand, advances in computing capabilities may be reaching the point of simulating the evolution of entire genome *in silico* (Dalquen et al., 2012); thus, creating biologically equivalence of digital organisms or artificial life. No surprises that the crucial link between field of biology and artificial life (Bersini, 2009) came from a bio-inspired computing model, genetic algorithms (GA), which modeled evolutionary processes in the computer. Recent studies had also suggested that artificial ecosystems exhibit characteristics that are similar to natural ecosystems (Dorin et al., 2008; Ronkko, 2007). In this study, the GA framework of Lim et al. (2010) was extended into Digital Organism Simulation Environment (DOSE) by providing an artificial ecosystem and a language to execute the DNA of the digital organisms.

Biological relevance is the main strength of DOSE. Each component and function is given a biological meaning. Most importantly, the genetic language, Ragaraja, is a 3-character instruction code and does not take on any operands. This is almost identical to the operations of natural DNA. However, there are two weaknesses of DOSE. Firstly, it is not able to simulate the 3-dimensional structure of protein molecules from Ragaraja code. Secondly, the current version of DOSE will require the user to be versed in Python programming as it is presented as a simulation library.

Future work in advancing DOSE may be the following – firstly, implementing a graphical user interface to ease construction of a simulation and to provide statistical tools to analyze results of the simulation. This will be useful to biologists using DOSE for their studies. Secondly, the execution of genome may be parallelized. This will allow DOSE to maximize the capabilities of multicore processors and cluster systems to increase simulation speed.

## Acknowledgement

The author wishes to thank Jack Oon (Immunology, University of Queensland, Australia) for his constructive comments.

## Appendix A: Ragaraja instruction set (Version 1)

The following table describes the action of each Ragaraja instructions. Of the 1000 possible instructions, 347 is defined, tested and used in this current version.

| Command | Description |
|---|---|
| 000 | Move forward by one cell on tape. Equivalent to ">" in BrainFuck. |
| 001 | Move forward by 5 cells on tape. Equivalent to 5 times of "000". |
| 002 | Move forward by 10 cells on tape. Equivalent to 10 times of "000". |
| 003 | Move forward by NxN cells on tape where N is the value of the current cell. If N is a decimal, it will move forward by the floor of NxN. For example, if N is 4.2, this operation will tape pointer forward by 17 cells. As NxN is always a positive number, it does not matter if the value of the current cell is positive or negative. |
| 004 | Move backward by one cell on tape. Equivalent to "<" in BrainFuck. |
| 005 | Move backward by 5 cells on tape. Equivalent to 5 times of "004". |
| 006 | Move backward by 10 cells on tape. Equivalent to 10 times of "004". |
| 007 | Move backward by NxN cells on tape where N is the value of the current cell. If N is a decimal, it will move backward by the floor of NxN. For example, if N is 4.2, this operation will tape pointer backward by 17 cells. |
| | As NxN is always a positive number, it does not matter if the value of the current cell is positive or negative. |
| 008 | Increase value of cell by 1. Equivalent to "+" in BrainFuck. |
| 009 | Increase value of cell by 5. Equivalent to 5 times of "008". |
| 010 | Increase value of cell by 10. Equivalent to 10 times of "008". |
| 011 | Decrease value of cell by 1. Equivalent to "-" in BrainFuck. |
| 012 | Decrease value of cell by 5. Equivalent to 5 times of "011". |
| 013 | Decrease value of cell by 10. Equivalent to 10 times of "011". |
| 016 | Add one cell to the end of the tape. |
| 017 | Add 10 cells to the end of the tape. |
| 018 | Remove one cell from the end of the tape. If original tape pointer is at the last cell before removal operation, the tape pointer will point to the last cell after removal. |
| 019 | Remove 10 cells from the end of the tape. If original tape pointer is at the last cell before removal operation, the tape pointer will point to the last cell after removal. |
| 020 | Output current tape cell value and append to the end of the output list. Equivalent to "." in BrainFuck. |
| 021 | Output current tape cell location and append to the end of the output list. |
| 022 | Output current source location and append to the end of the output list. |
| 023 | Move source pointer forward by one instruction without execution if the source pointer does not point beyond the length of the source after the move, otherwise, does not move the source pointer. |
| 024 | Move source pointer forward by 5 instructions without execution if the source pointer does not point beyond the length of the source after the move, otherwise, does not move the source pointer. |
| 025 | Move source pointer forward by 10 instructions without execution if the source pointer does not point beyond the length of the source after the move, otherwise, does not move the source pointer. |
| 032 | Double current tape cell value. |
| 033 | Half current tape cell value. |
| 034 | Insert a cell after the current tape cell. For example, if current tape cell is 35, a cell initialized to zero will be added as cell 36. As a result, the tape is 1 cell longer. |
| 035 | Delete the current cell. As a result, the tape is 1 cell shorter. |
| 036 | Delete the current and append to the end of the output list. As a result, the tape is 1 cell shorter. |
| 037 | Replace the current tape cell value with the last value of the output list, and delete the last value from the output list. |
| 038 | Replace the current tape cell value with the last value of the output list, without deleting the last value from the output list. |
| 039 | Replace the current tape cell value with the first value of the output list, and delete the first value from the output list. |
| 040 | Replace the current tape cell value with the first value of the output list, without deleting the first value from the output list. |
| 041 | Remove first value from the output list. |
| 042 | Remove last value from the output list. |
| 043 | Move the tape cell pointer to the first cell. |
| 044 | Move the tape cell pointer to the last cell. |
| 045 | Move the tape cell pointer to the location determined by the last value of the output list. If the last value of the output list is more than the length of the tape, it will take the modulus of the length of the tape. For example, the last value of the output list is 5, the tape cell pointer will point to the 5th cell on the tape. |
| 046 | Flip the tape. The original first cell becomes the last cell but the tape pointer does not flip in location. |
| 047 | Flip the output list. |
| 050 | Randomly execute "008" (increment by 1) or "000" (move forward by 1). |
| 051 | Randomly execute "011" (decrement by 1) or "004" (move backward by 1). |
| 052 | Randomly execute "000" (move forward by 1) or "004" (move backward by 1). |
| 053 | Randomly execute "008" (increment by 1) or "011" (decrement by 1). |
| 054 | Randomly execute "000" (move forward by 1) or "011" (decrement by 1). |
| 055 | Randomly execute "004" (move backward by 1) or "008" (increment by 1). |
| 056 | Randomly execute "000" (move forward by 1) or "004" (move backward by 1) or "011" (decrement by 1). |
| 057 | Randomly execute "000" (move forward by 1) or "008" (increment by 1) or "011" (decrement by 1). |
| 058 | Randomly execute "004" (move backward by 1) or "008" (Increment by 1) or "011" (decrement by 1). |
| 059 | Randomly execute "000" (move forward by 1) or "004" (move backward by 1) or "008" (increment by 1). |
| 060 | Randomly execute "000" (move forward by 1) or "004" (move backward by 1) or "008" (increment by 1) or "011" (decrement by 1). |
| 061 | Move forward by the number of cells signified by the current cell. |
| 062 | Move backward by the number of cells signified by the current cell. |
| 063 | Writes the first value of the input list into the current cell and remove the value from the input list. If input list |
| | is empty, "0" will be written. |
| 064 | Writes the first value of the input list into the current cell and without removing the value from the input list. If input list is empty, "0" will be written. |
| 065 | Add the value of the current cell (n) and (n+1)th cell, and store the value in the current cell. `Array[n] = Array[n] + Array[n+1]` |
| 066 | Add the value of the current cell (n) and first of the input list, and store the value in the current cell. |
| 067 | Add the value of the current cell (n) and last of the input list, and store the value in the current cell. |
| 068 | Subtract the value of the current cell (n) from (n+1)th cell, and store the value in the current cell. `Array[n] = Array[n+1] - Array[n]` |
| 069 | Subtract the value of the current cell (n) from the first of the input list, and store the value in the current cell. `Array[n] = InputList[0] - Array[n]` |
| 070 | Subtract the value of the current cell (n) from the last of the input list, and store the value in the current cell. `Array[n] = InputList[-1] - Array[n]` |
| 071 | Multiply the value of the current cell (n) and (n+1)th cell, and store the value in the current cell. `Array[n] = Array[n+1] * Array[n]` |
| 072 | Multiply the value of the current cell (n) and first of the input list, and store the value in the current cell. |
| 073 | Multiply the value of the current cell (n) and last of the input list, and store the value in the current cell. |
| 074 | Divide the value of the current cell (n) from (n+1)th cell, and store the value in the current cell. `Array[n] = Array[n+1] / Array[n]` |
| 075 | Divide the value of the current cell (n) from the first of the input list, and store the value in the current cell. `Array[n] = InputList[0] / Array[n]` |
| 076 | Divide the value of the current cell (n) from the last of the input list, and store the value in the current cell. `Array[n] = InputList[-1] - Array[n]` |
| 077 | Modulus (remainder after division) the value of the current cell (n) from (n+1)th cell, and store the value in the current cell. `Array[n] = Array[n+1] % Array[n]` |
| 078 | Modulus (remainder after division) the value of the current cell (n) from the first of the input list, and store the value in the current cell. `Array[n] = InputList[0] % Array[n]` |
| 079 | Modulus (remainder after division) the value of the current cell (n) from the last of the input list, and store the value in the current cell. `Array[n] = InputList[-1] % Array[n]` |
| 080 | Floor the value of the current cell. For example, if the value of the current cell is 6.7, it will become 6. |
| 081 | Swap the value of the current cell (n) and (n+1)th cell. |
| 084 | Set current tape cell to "0". |
| 085 | Set current tape cell to "-1". |
| 086 | Set current tape cell to "1". |
| 087 | Negate the value of the current cell. Positive value will be negative. Negative value will be positive. |
| 088 | Calculate the sine of the value of the current cell (measured in radians) and replace. `Array[n] = sine(Array[n])` |
| 089 | Calculate the cosine of the value of the current cell (measured in radians) and replace. `Array[n] = cosine(Array[n])` |
| 090 | Calculate the tangent of the value of the current cell (measured in radians) and replace. `Array[n] = tangent(Array[n])` |
| 091 | Calculate the arc sine of the value of the current cell (measured in radians) and replace. `Array[n] = arcsine(Array[n])` |
| 092 | Calculate the arc cosine of the value of the current cell (measured in radians) and replace. `Array[n] = arccosine(Array[n])` |
| 093 | Calculate the arc tangent of the value of the current cell (measured in radians) and replace. `Array[n] = arctangent(Array[n])` |
| 094 | Calculate the reciprocal of the value of the current cell (measured in radians) and replace. `Array[n] = 1/Array[n]` |
| 095 | Calculate the square root of the value of the current cell (measured in radians) and replace. `Array[n] = sqrt(Array[n])` |
| 096 | Calculate the natural logarithm of the value of the current cell (measured in radians) and replace. `Array[n] = ln(Array[n])` |
| 097 | Set the value of the current cell to pi (3.14159265358979323846) |
| 098 | Set the value of the current cell to e (2.718281828459045) |
| 099 | Calculate the hyperbolic sine of the value of the current cell (measured in radians) and replace. `Array[n] = sinh(Array[n])` |
| 100 | Calculate the hyperbolic cosine of the value of the current cell (measured in radians) and replace. `Array[n] = cosh(Array[n])` |
| 101 | Calculate the hyperbolic tangent of the value of the current cell (measured in radians) and replace. `Array[n] = tanh(Array[n])` |
| 102 | Calculate the hyperbolic arc sine of the value of the current cell (measured in radians) and replace. `Array[n] = arcsinh(Array[n])` |
| 103 | Calculate the hyperbolic arc cosine of the value of the current cell (measured in radians) and replace. `Array[n] = arccosh(Array[n])` |
| 104 | Calculate the hyperbolic arc tangent of the value of the current cell (measured in radians) and replace. `Array[n] = arctanh(Array[n])` |
| 105 | Convert the value of the current cell (measured in radians) to degrees and replace. |
| 106 | Convert the value of the current cell (measured in degrees) to radians and replace. |
| 107 | Raise the value of the current cell (n) to e, and store the value in the current cell. `Array[n] = Array[n]^e` |
| 108 | Raise e to the value of the current cell (n), and store the value in the current cell. `Array[n] = e^Array[n]` |
| 109 | Raise 10 to the value of the current cell (n), and store the value in the current cell. `Array[n] = 10^Array[n]` |
| 110 | Raise the value of the current cell (n) to the value of (n+1)th cell, and store the value in the current cell. `Array[n] = Array[n]^Array[n+1]` |
| 111 | Calculate the n-th root of the value of the current cell (n) where n is the value of (n+1)th cell, and store the value in the current cell. `Array[n] = Array[n]^(1/Array[n+1])` |
| 112 | Calculate the error function of the value of the current cell and replace. `Array[n] = erf(Array[n])` |
| 113 | Calculate the complementary error function of the value of the current cell and replace. `Array[n] = erfc(Array[n])` |
| 114 | Calculate the factorial of the integer value of the current cell (if the integer value is positive) and replace. `Array[n] = factorial(Array[n])` |
| 115 | Calculate the factorial of the absolute integer value of the current cell and replace. `Array[n] = factorial(abs(Array[n]))` |
| 116 | Calculate the Euclidean distance (hypotenuse) value of the current cell (n) to the value of (n+1)th cell, and store the value in the current cell. `Array[n] = sqrt(Array[n]*Array[n] + Array[n+1]*Array[n+1])` |
| 117 | Calculate the logarithm value of the current cell (n) to the base of the value of (n+1)th cell, and store the value in the current cell. `Array[n] = log(Array[n], base=Array[n+1])` |
| 120 | AND operator: Given positive numbers (>0) as True and zero or negative numbers (<=0)as False, store Array[current] AND Array[current+1] in the current cell (Array[current]) where "0" is False and "1" is True. |
| 121 | OR operator: Given positive numbers (>0) as True and zero or negative numbers (<=0)as False, store Array[current] OR Array[current+1] in the current cell (Array[current]) where "0" is False and "1" is True. |
| 122 | NOT operator: Given positive numbers (>0) as True and zero or negative numbers (<=0)as False, store NOT Array[current] in the current cell (Array[current]) where "0" is False and "1" is True. |
| 123 | LESS-THAN operator: Store Array[current] < Array[current+1] in the current cell (Array[current]) where "0" is False and "1" is True. |
| 124 | MORE-THAN operator: Store Array[current] > Array[current+1] in the current cell (Array[current]) where "0" is False and "1" is True. |
| 125 | EQUAL operator: Store Array[current] = Array[current+1] in the current cell (Array[current]) where "0" is False and "1" is True. |
| 126 | NOT-EQUAL operator: Store Array[current] != Array[current+1] in the current cell (Array[current]) where "0" is False and "1" is True. |
| 127 | LESS-THAN-OR-EQUAL operator: Store Array[current] <= Array[current+1] in the current cell (Array[current]) where "0" is False and "1" is True. |
| 128 | MORE-THAN-OR-EQUAL operator: Store Array[current] => Array[current+1] in the current cell (Array[current]) where "0" is False and "1" is True. |
| 140 | Move tape pointer to the centre of the tape. If the tape has odd number cells, it will move to the lower cell. For example, this instruction will move the tape pointer to the 500th cell of a 1000-cell tape, or 142nd of a 285-cell tape. |
| 141 | Move tape pointer to 1/4 the position of the tape. If the tape has odd number cells, it will move to the lower cell. For example, this instruction will move the tape pointer to the 250th cell of a 1000-cell tape, or 71st of a 285-cell tape. |
| 142 | Move tape pointer to 3/4 the position of the tape. If the tape has odd number cells, it will move to the lower cell. For example, this instruction will move the tape pointer to the 750th cell of a 1000-cell tape, or 213rd of a 285-cell tape. |
| 143 | Move tape pointer to the position as the integer value in the current cell. If the value of the cell is larger than the length of the tape, it will move to the modulus of the integer value in the current cell. |
| 144 | Divide current cell value by 10. |
| 145 | Multiply current cell value by 10. |
| 146 | Add all cell values from (n+1)th cell to the end of the tape and store result in current cell (n).<br>`Array[n] = sum(Array[n+1:])` |
| 147 | Add all cell values from n-th cell to the end of the tape and store result in current cell (n).<br>`Array[n] = sum(Array[n:])` |
| 150 | Add all cell values in the tape and store result in current cell (n). `Array[n] = sum(Array[:])` |
| 151 | Average all cell values from (n+1)th cell to the end of the tape and store result in current cell (n).<br>`Array[n] = average(Array[n+1:])` |
| 152 | Average all cell values from n-th cell to the end of the tape and store result in current cell (n).<br>`Array[n] = average(Array[n:])` |
| 153 | Average all cell values from first cell to the cell before n-th cell and store result in current cell (n).<br>`Array[n] = average(Array[0:n])` |
| 154 | Average all cell values from first cell to n-th cell (inclusive) and store result in current cell (n).<br>`Array[n] = average(Array[0:n+1])` |
| 155 | Half every cell value in tape. |
| 156 | Double every cell value in tape. |
| 157 | Divide every cell value in tape by 10. |
| 158 | Multiply every cell value in tape by 10. |
| 159 | Divide every cell value in tape by 100. |
| 160 | Multiply every cell value in tape by 100. |
| 161 | Cut the tape before the current cell (n) and append it to the end of the tape and set tape pointer to 0.<br>`<---A--->n<---B--->` ➔ `n<---B---><---A--->` |
| 162 | Cut the tape after the current cell (n) and append it to the start of the tape and set tape pointer to the last cell.<br>`<---A--->n<---B--->` ➔ `<---B---><---A--->n` |
| 163 | Cut out the current cell and append it to the front of the tape and set tape pointer to 0.<br>`<---A--->n<---B--->` ➔ `n<---A---><---B--->` |
| 164 | Cut out the current cell and append it to the end of the tape and set tape pointer to the last cell.<br>`<---A--->n<---B--->` ➔ `<---A---><---B--->n` |
| 165 | Multiply every cell value in tape by -1. |
| 166 | Square all the cell values in the cells after the current cell (current cell value is not affected). |
| 167 | Square all the cell values in the cells before the current cell (current cell value is not affected). |
| 168 | Square every cell value in tape. |
| 169 | Square root every cell value in tape. |
| 170 | Square root all the cell values in the cells after the current cell (current cell value is not affected). |
| 171 | Square root all the cell values in the cells before the current cell (current cell value is not affected). |
| 189 | Set all values in tape to "0". |
| 196 | Set the value of the current cell to the standard deviation of the values in the tape. |
| 197 | Set the value of the current cell to the geometric mean of the values in the tape. |
| 198 | Set the value of the current cell to the harmonic mean of the values in the tape. |
| 201 | Store value of current tape cell to register #1 |
| 202 | Store value of current tape cell to register #2 |
| 203 | Store value of current tape cell to register #3 |
| 204 | Store value of current tape cell to register #4 |
| 205 | Store value of current tape cell to register #5 |
| 206 | Store value of current tape cell to register #6 |
| 207 | Store value of current tape cell to register #7 |
| 208 | Store value of current tape cell to register #8 |
| 209 | Store value of current tape cell to register #9 |
| 210 | Store value of current tape cell to register #10 |
| 211 | Store value of current tape cell to register #11 |
| 212 | Store value of current tape cell to register #12 |
| 213 | Store value of current tape cell to register #13 |
| 214 | Store value of current tape cell to register #14 |
| 215 | Store value of current tape cell to register #15 |
| 216 | Store value of current tape cell to register #16 |
| 217 | Store value of current tape cell to register #17 |
| 218 | Store value of current tape cell to register #18 |
| 219 | Store value of current tape cell to register #19 |
| 220 | Store value of current tape cell to register #20 |
| 221 | Store value of current tape cell to register #21 |
| 222 | Store value of current tape cell to register #22 |
| 223 | Store value of current tape cell to register #23 |
| 224 | Store value of current tape cell to register #24 |
| 225 | Store value of current tape cell to register #25 |
| 226 | Store value of current tape cell to register #26 |
| 227 | Store value of current tape cell to register #27 |
| 228 | Store value of current tape cell to register #28 |
| 229 | Store value of current tape cell to register #29 |
| 230 | Store value of current tape cell to register #30 |
| 231 | Store value of current tape cell to register #31 |
| 232 | Store value of current tape cell to register #32 |
| 233 | Store value of current tape cell to register #33 |
| 234 | Store value of current tape cell to register #34 |
| 235 | Store value of current tape cell to register #35 |
| 236 | Store value of current tape cell to register #36 |
| 237 | Store value of current tape cell to register #37 |
| 238 | Store value of current tape cell to register #38 |
| 239 | Store value of current tape cell to register #39 |
| 240 | Store value of current tape cell to register #40 |
| 241 | Store value of current tape cell to register #41 |
| 242 | Store value of current tape cell to register #42 |
| 243 | Store value of current tape cell to register #43 |
| 244 | Store value of current tape cell to register #44 |
| 245 | Store value of current tape cell to register #45 |
| 246 | Store value of current tape cell to register #46 |
| 247 | Store value of current tape cell to register #47 |
| 248 | Store value of current tape cell to register #48 |
| 249 | Store value of current tape cell to register #49 |
| 250 | Store value of current tape cell to register #50 |
| 251 | Store value of current tape cell to register #51 |
| 252 | Store value of current tape cell to register #52 |
| 253 | Store value of current tape cell to register #53 |
| 254 | Store value of current tape cell to register #54 |
| 255 | Store value of current tape cell to register #55 |
| 256 | Store value of current tape cell to register #56 |
| 257 | Store value of current tape cell to register #57 |
| 258 | Store value of current tape cell to register #58 |
| 259 | Store value of current tape cell to register #59 |
| 260 | Store value of current tape cell to register #60 |
| 261 | Store value of current tape cell to register #61 |
| 262 | Store value of current tape cell to register #62 |
| 263 | Store value of current tape cell to register #63 |
| 264 | Store value of current tape cell to register #64 |
| 265 | Store value of current tape cell to register #64 |
| 266 | Store value of current tape cell to register #66 |
| 267 | Store value of current tape cell to register #67 |
| 268 | Store value of current tape cell to register #68 |
| 269 | Store value of current tape cell to register #69 |
| 270 | Store value of current tape cell to register #70 |
| 271 | Store value of current tape cell to register #71 |
| 272 | Store value of current tape cell to register #72 |
| 273 | Store value of current tape cell to register #73 |
| 274 | Store value of current tape cell to register #74 |
| 275 | Store value of current tape cell to register #75 |
| 276 | Store value of current tape cell to register #76 |
| 277 | Store value of current tape cell to register #77 |
| 278 | Store value of current tape cell to register #78 |
| 279 | Store value of current tape cell to register #79 |
| 280 | Store value of current tape cell to register #80 |
| 281 | Store value of current tape cell to register #81 |
| 282 | Store value of current tape cell to register #82 |
| 283 | Store value of current tape cell to register #83 |
| 284 | Store value of current tape cell to register #84 |
| 285 | Store value of current tape cell to register #85 |
| 286 | Store value of current tape cell to register #86 |
| 287 | Store value of current tape cell to register #87 |
| 288 | Store value of current tape cell to register #88 |
| 289 | Store value of current tape cell to register #89 |
| 290 | Store value of current tape cell to register #90 |
| 291 | Store value of current tape cell to register #91 |
| 292 | Store value of current tape cell to register #92 |
| 293 | Store value of current tape cell to register #93 |
| 294 | Store value of current tape cell to register #94 |
| 295 | Store value of current tape cell to register #95 |
| 296 | Store value of current tape cell to register #96 |
| 297 | Store value of current tape cell to register #97 |
| 298 | Store value of current tape cell to register #98 |
| 299 | Store value of current tape cell to register #99 |
| 301 | Put value from register #1 to current tape cell |
| 302 | Put value from register #2 to current tape cell |
| 303 | Put value from register #3 to current tape cell |
| 304 | Put value from register #4 to current tape cell |
| 305 | Put value from register #5 to current tape cell |
| 306 | Put value from register #6 to current tape cell |
| 307 | Put value from register #7 to current tape cell |
| 308 | Put value from register #8 to current tape cell |
| 309 | Put value from register #9 to current tape cell |
| 310 | Put value from register #10 to current tape cell |
| 311 | Put value from register #11 to current tape cell |
| 312 | Put value from register #12 to current tape cell |
| 313 | Put value from register #13 to current tape cell |
| 314 | Put value from register #14 to current tape cell |
| 315 | Put value from register #15 to current tape cell |
| 316 | Put value from register #16 to current tape cell |
| 317 | Put value from register #17 to current tape cell |
| 318 | Put value from register #18 to current tape cell |
| 319 | Put value from register #19 to current tape cell |
| 320 | Put value from register #20 to current tape cell |
| 321 | Put value from register #21 to current tape cell |
| 322 | Put value from register #22 to current tape cell |
| 323 | Put value from register #23 to current tape cell |
| 324 | Put value from register #24 to current tape cell |
| 325 | Put value from register #25 to current tape cell |
| 326 | Put value from register #26 to current tape cell |
| 327 | Put value from register #27 to current tape cell |
| 328 | Put value from register #28 to current tape cell |
| 329 | Put value from register #29 to current tape cell |
| 330 | Put value from register #30 to current tape cell |
| 331 | Put value from register #31 to current tape cell |
| 332 | Put value from register #32 to current tape cell |
| 333 | Put value from register #33 to current tape cell |
| 334 | Put value from register #34 to current tape cell |
| 335 | Put value from register #35 to current tape cell |
| 336 | Put value from register #36 to current tape cell |
| 337 | Put value from register #37 to current tape cell |
| 338 | Put value from register #38 to current tape cell |
| 339 | Put value from register #39 to current tape cell |
| 340 | Put value from register #40 to current tape cell |
| 341 | Put value from register #41 to current tape cell |
| 342 | Put value from register #42 to current tape cell |
| 343 | Put value from register #43 to current tape cell |
| 344 | Put value from register #44 to current tape cell |
| 345 | Put value from register #45 to current tape cell |
| 346 | Put value from register #46 to current tape cell |
| 347 | Put value from register #47 to current tape cell |
| 348 | Put value from register #48 to current tape cell |
| 349 | Put value from register #49 to current tape cell |
| 350 | Put value from register #50 to current tape cell |
| 351 | Put value from register #51 to current tape cell |
| 352 | Put value from register #52 to current tape cell |
| 353 | Put value from register #53 to current tape cell |
| 354 | Put value from register #54 to current tape cell |
| 355 | Put value from register #55 to current tape cell |
| 356 | Put value from register #56 to current tape cell |
| 357 | Put value from register #57 to current tape cell |
| 358 | Put value from register #58 to current tape cell |
| 359 | Put value from register #59 to current tape cell |
| 360 | Put value from register #60 to current tape cell |
| 361 | Put value from register #61 to current tape cell |
| 362 | Put value from register #62 to current tape cell |
| 363 | Put value from register #63 to current tape cell |
| 364 | Put value from register #64 to current tape cell |
| 365 | Put value from register #65 to current tape cell |
| 366 | Put value from register #66 to current tape cell |
| 367 | Put value from register #67 to current tape cell |
| 368 | Put value from register #68 to current tape cell |
| 369 | Put value from register #69 to current tape cell |
| 370 | Put value from register #70 to current tape cell |
| 371 | Put value from register #71 to current tape cell |
| 372 | Put value from register #72 to current tape cell |
| 373 | Put value from register #73 to current tape cell |
| 374 | Put value from register #74 to current tape cell |
| 375 | Put value from register #75 to current tape cell |
| 376 | Put value from register #76 to current tape cell |
| 377 | Put value from register #77 to current tape cell |
| 378 | Put value from register #78 to current tape cell |
| 379 | Put value from register #79 to current tape cell |
| 380 | Put value from register #80 to current tape cell |
| 381 | Put value from register #81 to current tape cell |
| 382 | Put value from register #82 to current tape cell |
| 383 | Put value from register #83 to current tape cell |
| 384 | Put value from register #84 to current tape cell |
| 385 | Put value from register #85 to current tape cell |
| 386 | Put value from register #86 to current tape cell |
| 387 | Put value from register #87 to current tape cell |
| 388 | Put value from register #88 to current tape cell |
| 389 | Put value from register #89 to current tape cell |
| 390 | Put value from register #90 to current tape cell |
| 391 | Put value from register #91 to current tape cell |
| 392 | Put value from register #92 to current tape cell |
| 393 | Put value from register #93 to current tape cell |
| 394 | Put value from register #94 to current tape cell |
| 395 | Put value from register #95 to current tape cell |
| 396 | Put value from register #96 to current tape cell |
| 397 | Put value from register #97 to current tape cell |
| 398 | Put value from register #98 to current tape cell |
| 399 | Put value from register #99 to current tape cell |

For direct simulation of naturally occurring DNA sequences, a subset of Ragaraja is defined based on IUPAC nucleotide codes, known as NucleotideBF[2], which can be set in DOSE parameters as Ragaraja version 0.1. Guanine (G) and cytosine (C) are paired to move left and right of the tape respectively, while adenine (A) and thymine (T) are paired for increment and decrement respectively. Alternatively, user can define their set of instruction usage by toggling each instruction in *ragaraja_instructions.txt* file, "Y" for use and "N" for not used, and setting Ragaraja version as 0.

| **Instructions** | | | **Description** |
|---|---|---|---|
| **NucleotideBF** | **Ragaraja** | **BrainFuck** | |
| G | 000 | > | Move the pointer to the right. |
| C | 004 | < | Move the pointer to the left. |
| A | 008 | + | Increment the memory cell under the pointer. |
| T | 011 | - | Decrement the memory cell under the pointer. |
| . | 020 | . | Full stop. Output the character signified by the cell at the pointer. |
| R | 050 | | Random between A or G. |
| Y | 051 | | Random between C or T. |
| S | 052 | | Random between G or C. |
| W | 053 | | Random between A or T. |
| K | 054 | | Random between G or T. |
| M | 055 | | Random between A or C. |
| B | 056 | | Random between C or G or T. |
| D | 057 | | Random between A or G or T. |
| H | 058 | | Random between A or C or T. |
| V | 059 | | Random between A or C or G. |
| N | 060 | | Random between A or T or C or G. |

## Appendix B: DOSE world code (file name: *dose_world.py*)

```
'''
World structure for DOSE (digital organism simulation environment)
Date created: 13th September 2012
Licence: Python Software Foundation License version 2
'''
import copy
import cPickle

class World(object):
    '''
    Representation of a 3-dimensional ecological world.

    The ecosystem is made up of ecological cells. Each ecological cell is
    modelled as a dictionary of
     -  local_input: A list containing processed input, representing the
        partial local ecological condition, to be used as input to the organisms in
        the current ecological cell. This is updated by World.update_local
        function.
     -  local_output: A list containing processed output, representing the
        partial local ecological condition. This is updated by World.update_local
        function.
     -  temporary_input: A list acting as temporary holding for input after
        being fed to the organisms in the current ecological cell, which is to be
        used to update local_input and local_output lists by World.update_local and
        World.update_ecology functions.
     -  temporary_output: A list acting as temporary holding for output from the
```

[2] http://esolangs.org/wiki/NucleotideBF_(nBF)

```
        organisms in the current ecological cell, which is to be used to update
        local_input and local_output lists by World.update_local and
        World.update_ecology functions.
     -  organisms: The number of organisms in the current ecological cell
        which is updated by World.organism_movement and World.organism_location
        functions.
    '''
    ecosystem = {}

    def __init__(self, world_x, world_y, world_z):
        '''
        Setting up the world and ecosystem

        @param world_x: number of ecological cells on the x-axis
        @type world_x: integer
        @param world_y: number of ecological cells on the y-axis
        @type world_y: integer
        @param world_z: number of ecological cells on the z-axis
        @type world_z: integer
        '''
        eco_cell = {'local_input': [], 'local_output': [],
                    'temporary_input': [], 'temporary_output': [],
                    'organisms': 0}
        self.world_x = int(world_x)
        self.world_y = int(world_y)
        self.world_z = int(world_z)
        for x in range(self.world_x):
            eco_x = {}
            for y in range(self.world_y):
                eco_y = {}
                for z in range(self.world_z):
                    eco_y[z] = copy.deepcopy(eco_cell)
                eco_x[y] = copy.deepcopy(eco_y)
            self.ecosystem[x] = copy.deepcopy(eco_x)

    def eco_burial(self, filename):
        '''
        Function to preserve the entire ecosystem.

        @param filename: file name of preserved ecosystem.
        '''
        f = open(filename, 'w')
        cPickle.dump(self.ecosystem, f)
        f.close()

    def eco_excavate(self, filename):
        '''
        Function to excavate entire ecosystem.

        @param filename: file name of preserved ecosystem.
        '''
        self.ecosystem = cPickle.load(open(filename, 'r'))

    def ecoregulate(self):
        '''
        Function to simulate events to the entire ecosystem. B{This
        function may be over-ridden by the inherited class or substituted
        to cater for ecological schemes but not an absolute requirement to
        do so.}
        '''
        pass

    def organism_movement(self, x, y, z):
        '''
```

```
        Function to trigger organism movement from current ecological cell
        to an adjacent ecological cell. B{This function may be over-ridden
        by the inherited class or substituted to cater for mobility
        schemes but not an absolute requirement to do so.}

        @param x: location of current ecological cell on the x-axis
        @type x: integer
        @param y: location of current ecological cell on the y-axis
        @type y: integer
        @param z: location of current ecological cell on the z-axis
        @type z: integer
        '''
        pass
    def organism_location(self, x, y, z):
        '''
        Function to trigger organism movement from current ecological cell
        to a distant ecological cell. B{This function may be over-ridden
        by the inherited class or substituted to cater for mobility
        schemes but not an absolute requirement to do so.}

        @param x: location of current ecological cell on the x-axis
        @type x: integer
        @param y: location of current ecological cell on the y-axis
        @type y: integer
        @param z: location of current ecological cell on the z-axis
        @type z: integer
        '''
        pass

    def update_ecology(self, x, y, z):
        '''
        Function to process temporary_input and temporary_output from the
        activities of the organisms in the current ecological cell into a
        local ecological cell condition, and update the ecosystem.
        B{This function may be over-ridden by the inherited class or
        substituted to cater for ecological schemes but not an absolute
        requirement to do so.}

        @param x: location of current ecological cell on the x-axis
        @type x: integer
        @param y: location of current ecological cell on the y-axis
        @type y: integer
        @param z: location of current ecological cell on the z-axis
        @type z: integer
        '''
        pass

    def update_local(self, x, y, z):
        '''
        Function to update local ecological cell condition from the
        ecosystem.
        B{This function may be over-ridden by the inherited class or
        Substituted to cater for ecological schemes but not an absolute
        requirement to do so.}

        @param x: location of current ecological cell on the x-axis
        @type x: integer
        @param y: location of current ecological cell on the y-axis
        @type y: integer
        @param z: location of current ecological cell on the z-axis
        @type z: integer
        '''
        pass
```

```
    def report(self):
        '''
        Function to report the status of the world and ecosystem. B{This
        function may be over-ridden by the inherited class or substituted
        to cater for specific reporting schemes but not an absolute
        requirement to do so.}

        @return: dictionary of status describing the current generation
        '''
        pass
```

## Appendix C: Boilerplate codes for DOSE simulation

A DOSE simulation consists of 2 main files – *dose_parameters.py*, which contains the essential parameters needed for the simulation, and *dose_entities.py*, which contains the inherited classes to over-ride the 12 user-defined functions listed in Table 1. These files will be used by run_dose.py, which contains the default simulation driver.

### File name: dose_parameters.py

```
'''
Initial chromosome (list) for the ancestor organism
'''
initial_chromosome = ['0', '0', '0', '0', '0', '0', '0', '0', '0', '0', '0',
'0', '0', '0', '0', '0', '0', '0', '0', '0', '0', '0', '0', '0', '0', '0',
'0', '0', '0', '0', '0', '0', '0', '0', '0', '0', '0', '0', '0', '0', '0',
'0', '0', '0', '0', '0', '0', '0', '0', '0', '0', '0', '0', '0', '0', '0',
'0', '0', '0', '0']

'''
Probability of number of mutations per base. For example, 0.1 means 10% of
the chromosome will be mutated per generation.
'''
background_mutation_rate = 0.1

'''
Probability of mutation per base above additional_mutation_rate. No mutation
event will ever happen if (additional_mutation_rate + additional_mutation_rate)
is less than zero.
'''
additional_mutation_rate = 0

'''
Size of cytoplasm (length of list) for the ancestor organism.
'''
cytoplasm_size = 50

'''
Maximum size of cytoplasm (length of list) allowable. Some Ragaraja instructions
can increase cytoplasms size, equivalent to cell enlargement.
'''
max_cytoplasm_size = 200

'''
If 'True', a clean cytoplasm will be used to execute the genome. If 'False',
the cytoplasm will be reused in the next generation.
'''
clean_cytoplasm = True

'''
Maximum number of codons to evaluate. Therfore, the maximum evaluated genome
```

```
is 3*max_codon bases.
'''
max_codon = 2000

'''
List containing name(s) of population. The number of names = the number of
populations to create.
'''
population_names = ['pop1', 'pop2']

'''
Number of organism per population.
'''
population_size = 100

'''
Number of ecological cells in the world in (x,y,z) coordinate as
(world_x, world_y, world_z).
'''
world_x = 5
world_y = 5
world_z = 5

'''
List of tuple indicating the location on ecosystem to map the population(s) and
will be mapped in the same order as population_names. For example, if
population_names = ['pop1', 'pop2'] and
population_locations = [(0,0,0), (4,4,4)],
population 'pop1' will be in (0,0,0) and population 'pop2' will be in (4,4,4).
'''
population_locations = [(0,0,0), (4,4,4)]

'''
Maximum number of generations to simulate.
'''
maximum_generations = 500

'''
Number of generations between each freezing/fossilization event.
Freezing/fossilization event is similar to glycerol stocking in microbiology.
'''
fossilized_frequency = 100

'''
Proportion of population to freeze/fossilize. If the population size or the
preserved proportion is below 100, the entire population will be preserved.
'''
fossilized_ratio = 0.01

'''
Dictionary containing prefix of file names for freezing/fossilization. The
preserved sample will be written into a file with name in the following
format - <prefix>_<generation count>_<sample size>.gap
'''
fossil_files = {'pop1': 'pop1', 'pop2': 'pop2'}

'''
Number of generations between printing of reports (based on Population.report
function) into files. The result file format is <UTC date time stamp>
<current generation count> <output from Population.report function>.
'''
print_frequency = 10

'''
```

```
Dictionary containing file names for result files. File name will be
<file names>.result
'''
result_files = {'pop1': 'pop1', 'pop2': 'pop2'}

'''
Version of Ragaraja instruction set to use.
'''
ragaraja_version = 0.1

'''
File containing Ragaraja instruction set to be used. This option is only
effective when ragaraja_version = 0. Format of file is <instruction>={Y|N}
where "Y" = instruction to be used and "N" = instruction not to be used.
'''
user_defined_instructions = 'ragaraja_instructions.txt'

'''
Number of generations between each burial/preservation of ecosystem.
'''
eco_buried_frequency = 500

'''
Prefix of file name for ecosystem burial/preservation. The preserved ecosystem
will be written into a file with name in the following format -
<prefix>_<generation count>.eco
'''
eco_burial_file = 'eco'
```

## File name: dose_entities.py

```
'''
Boiler-plate codes for DOSE (digital organism simulation environment) entities
Date created: 13th September 2012
Licence: Python Software Foundation License version 2
'''

import genetic as g
import dose_world as w
from dose_parameters import initial_chromosome, background_mutation_rate
from dose_parameters import cytoplasm_size, population_size
from dose_parameters import maximum_generations
from dose_parameters import world_x, world_y, world_z

Chromosome = g.Chromosome(initial_chromosome,
                          ['0','1','2','3','4','5','6','7','8','9'],
                          background_mutation_rate)

class Organism(g.Organism):

    cytoplasm = [0]*cytoplasm_size

    def __init__(self): self.genome = [Chromosome.replicate()]
    def get_cytoplasm(self):
        return ','.join([str(x) for x in self.cytoplasm])
    def fitness(self): pass
    def mutation_scheme(self): pass

class Population(g.Population):

    def __init__(self, pop_size=population_size,
                 max_gen=maximum_generations):
        self.agents = [Organism() for x in xrange(pop_size)]
        self.generation = 0
```

```
        self.maximum_generations = max_gen
    def prepopulation_control(self): pass
    def mating(self): pass
    def postpopulation_control(self): pass
    def generation_events(self): pass
    def report(self): pass

class World(w.World):
    def __init__(self, world_x=world_x, world_y=world_y, world_z=world_z):
        super(World, self).__init__(world_x, world_y, world_z)
    def organism_movement(self, x, y, z): pass
    def organism_location(self, x, y, z): pass
    def ecoregulate(self): pass
    def update_ecology(self, x, y, z): pass
    def update_local(self, x, y, z): pass
    def report(self): pass
```

## Appendix D: Default simulation driver (file name: *run_dose.py*)

```
import sys
from datetime import datetime
import ragaraja as N
import register_machine as r
from dose_parameters import *

# Set Ragaraja instruction version
if ragaraja_version == 0:
    f = open(user_defined_instructions, 'r').readlines()
    f = [x[:-1].split('=') for x in f]
    f = [x[0] for x in f if x[1] == 'Y']
    ragaraja_instructions = f
if ragaraja_version == 0.1:
    ragaraja_instructions = N.nBF_instructions
if ragaraja_version == 1:
    ragaraja_instructions = N.ragaraja_v1
for instruction in N.ragaraja:
    if instruction not in ragaraja_instructions:
        N.ragaraja[instruction] = N.not_used

# Write DOSE parameters into result files
for name in population_names:
    f = open(result_files[name] + '.result.txt', 'a')
    f.write('STARTING SIMULATION - ' + str(datetime.utcnow()) + '\n')
    f.write('DOSE parameters:' + '\n')
    f.write('initial_chromosome = ' + str(initial_chromosome) + '\n')
    f.write('chromosome_size = ' + str(len(initial_chromosome)) + '\n')
    f.write('cytoplasm_size = ' + str(cytoplasm_size) + '\n')
    f.write('population_size = ' + str(population_size) + '\n')
    f.write('population_names = ' + str(population_names) + '\n')
    f.write('world_x = ' + str(world_x) + '\n')
    f.write('world_y = ' + str(world_y) + '\n')
    f.write('world_z = ' + str(world_z) + '\n')
    f.write('population_locations = ' + str(population_locations) + '\n')
    f.write('background_mutation_rate = ' + str(background_mutation_rate) + '\n')
    f.write('additional_mutation_rate = ' + str(additional_mutation_rate) + '\n')
    f.write('maximum_generations = ' + str(maximum_generations) + '\n')
    f.write('fossilized_ratio = ' + str(fossilized_ratio) + '\n')
    f.write('fossilized_frequency = ' + str(fossilized_frequency) + '\n')
    f.write('fossil_files = ' + str(fossil_files) + '\n')
    f.write('print_frequency = ' + str(print_frequency) + '\n')
    f.write('result_files = ' + str(result_files) + '\n')
    f.write('ragaraja_version = ' + str(ragaraja_version) + '\n')
```

```
    f.write('instruction_set = ' + str(ragaraja_instructions) + '\n')
    f.close()

def simulate(entity_module):
    exec('from %s import World, Population' % entity_module)

    populations = {}
    world = World()

    for i in range(len(population_names)):
        populations[population_names[i]] = Population()
        L = population_locations[i]
        world.ecosystem[L[0]][L[1]][L[2]]['organisms'] = \
            len(populations[population_names[i]].agents)
        for x in range(len(populations[population_names[i]].agents)):
            populations[population_names[i]].agents[x].status['location'] = L

    ####################################################################
    # Default Simulation Driver                                        #
    # (do not change anything above this line)                         #
    ####################################################################
    generation_count = 0
    while generation_count < maximum_generations:
        generation_count = generation_count + 1
        '''
        Run World.ecoregulate function
        '''
        world.ecoregulate()

        '''
        For each ecological cell, run World.update_ecology and
        World.update_local functions
        '''
        for x in range(world.world_x):
            for y in range(world.world_y):
                for z in range(world.world_z):
                    world.update_ecology(x, y, z)
                    world.update_local(x, y, z)

        '''
        For each organism
            Execute genome by Ragaraja interpreter using
               existing cytoplasm, local conditions as input
            Update cytoplasm (Organism.cytoplasm)
            Add input/output from organism intermediate condition of local cell
        '''
        for name in population_names:
            for i in range(len(populations[name].agents)):
                source = populations[name].agents[i].genome[0].sequence
                source = ''.join(source)
                if clean_cytoplasm:
                    array = [0]*cytoplasm_size
                else:
                    array = populations[name].agents[i].cytoplasm
                L = populations[name].agents[i].status['location']
                inputdata = world.ecosystem[L[0]][L[1]][L[2]]['local_input']
                try: (array, apointer, inputdata,
                      output, source, spointer) = \
                        r.interpret(source, N.ragaraja, 3,
                                    inputdata, array,
                                    max_cytoplasm_size,
                                    max_codon)
                except IndexError: pass
                except ZeroDivisionError: pass
```

```
                except OverflowError: pass
                except ValueError: pass
                populations[name].agents[i].cytoplasm = array
                world.ecosystem[L[0]][L[1]][L[2]]['temporary_input'] = inputdata
                world.ecosystem[L[0]][L[1]][L[2]]['temporary_output'] = output

        '''
        For each population
            Run Population.prepopulation_control function
            Run Population.mating function and add new organisms to cell
            For each organism, run Organism.mutation_scheme function
            Run Population.generation_events function
            Add 1 to generation count
            Run Population.report function
            Fossilize population if needed
        '''
        for name in population_names:
            report = populations[name].generation_step()
            if generation_count % int(fossilized_frequency) == 0:
                ffile = fossil_files[name] + '_'
                populations[name].freeze(ffile, fossilized_ratio)
            if generation_count % int(print_frequency) == 0:
                print str(generation_count), str(report)
                f = open(result_files[name] + '.result.txt', 'a')
                dtstamp = str(datetime.utcnow())
                f.write('\t'.join([dtstamp, str(generation_count),
                                   str(report)]))
                f.write('\n')
                f.close()

        '''
        For each ecological cell
            Run World.organism_movement function
            Run World.organism_location function
            Run World.report function
        '''
        for x in range(world.world_x):
            for y in range(world.world_y):
                for z in range(world.world_z):
                    world.organism_movement(x, y, z)
                    world.organism_location(x, y, z)
                    world.report()

        '''
        Bury ecosystem if needed
        '''
        if generation_count % int(eco_buried_frequency) == 0:
            filename = eco_burial_file + '_' + str(generation_count) + '.eco'
            world.eco_burial(filename)

if __name__ == "__main__":
    entity_module = sys.argv[1]
    simulate(entity_module)
```

## Appendix E: DOSE entities used for experiment

```
import random
import genetic as g
import dose_world as w
from dose_parameters import initial_chromosome, background_mutation_rate
from dose_parameters import cytoplasm_size, population_size
from dose_parameters import maximum_generations
```

```
from dose_parameters import world_x, world_y, world_z

Chromosome = g.Chromosome(initial_chromosome,
                          ['0','1','2','3','4','5','6','7','8','9'],
                          background_mutation_rate)

class Organism(g.Organism):

    cytoplasm = [0]*cytoplasm_size

    def __init__(self): self.genome = [Chromosome.replicate()]
    def get_cytoplasm(self):
        return ','.join([str(x) for x in self.cytoplasm])
    def fitness(self): pass
    def mutation_scheme(self):
        self.genome[0].rmutate('point', 0)

class Population(g.Population):

    def __init__(self, pop_size=population_size,
                 max_gen=maximum_generations):
        self.agents = [Organism() for x in xrange(pop_size)]
        self.generation = 0
        self.maximum_generations = max_gen
    def prepopulation_control(self): pass
    def mating(self): pass
    def postpopulation_control(self): pass
    def generation_events(self): pass
    def report(self):
        sequences = [''.join(org.genome[0].sequence)
                     for org in self.agents]
        return '\t'.join(sequences)

class World(w.World):
    def __init__(self, world_x=world_x, world_y=world_y, world_z=world_z):
        super(World, self).__init__(world_x, world_y, world_z)
    def organism_movement(self, x, y, z): pass
    def organism_location(self, x, y, z): pass
    def ecoregulate(self): pass
    def update_ecology(self, x, y, z): pass
    def update_local(self, x, y, z): pass
    def report(self): pass
```